\pdfoutput=1

\documentclass[11pt]{article}

\usepackage{acl}

\usepackage{times}
\usepackage{latexsym}
\usepackage{courier}
\usepackage[T1]{fontenc}

\usepackage[utf8]{inputenc}

\usepackage{microtype}

\usepackage{inconsolata}

\usepackage{tabularx}
\usepackage{times}
\usepackage{latexsym}
\usepackage{hyperref}
\usepackage{url}
\usepackage{bbm}
\usepackage{amsmath,amsfonts,bm}
\usepackage{graphicx}
\usepackage{multirow}
\usepackage{makecell}
\usepackage{array}
\usepackage[para]{threeparttable}
\usepackage{booktabs}
\usepackage{xspace}
\usepackage{subcaption}
\usepackage{multirow}
\usepackage[normalem]{ulem}
\usepackage{float}
\usepackage{afterpage}
\usepackage{dirtytalk}
\usepackage{bbding}
\usepackage{amssymb}

\newcommand{\method}{\textsc{LangBridge}\xspace}
\newcommand{\methodtable}{\textbf{LB}\xspace}

\newcommand{\blue}{\textcolor[HTML]{0066CC}}
\newcolumntype{s}{>{\hsize=.5\hsize}X}

\title{\method: Multilingual Reasoning Without Multilingual Supervision}

\author{Dongkeun Yoon{\textsuperscript{1}}\quad Joel Jang{\textsuperscript{2}}\quad Sungdong Kim{\textsuperscript{1, 3}} \\ \textbf{Seungone Kim{\textsuperscript{1, 4}}}\quad 
\textbf{Sheikh Shafayat{\textsuperscript{1}}}  \quad \textbf{Minjoon Seo{\textsuperscript{1}}} \\ \\
{\textsuperscript{1}}KAIST \quad{\textsuperscript{2}} University of Washington \quad{\textsuperscript{3}} NAVER AI Lab \quad{\textsuperscript{4}} Carnegie Mellon University \\
\texttt{\{dkyoon, minjoon\}@kaist.ac.kr} \\ 
}

\begin{document}
\maketitle
\begin{abstract}
 
We introduce \method, a \textit{zero-shot} approach to adapt language models for multilingual reasoning tasks without multilingual supervision. \method operates by \say{bridging} two models, each specialized in different aspects: (1) one specialized in understanding multiple languages (e.g., mT5 encoder) and (2) one specialized in reasoning (e.g., Orca 2). \method connects the two models by introducing minimal trainable parameters between them. Despite utilizing only English data for training, \method considerably enhances the performance of language models on low-resource languages across mathematical reasoning, code completion, logical reasoning, and commonsense reasoning. Our analysis suggests that the efficacy of \method stems from the language-agnostic characteristics of multilingual representations. 
We publicly release our code and models.\footnote{\href{https://github.com/kaistAI/LangBridge}{github.com/kaistAI/LangBridge}}

\end{abstract}

\section{Introduction}
Language models (LMs) are known to exhibit inferior performance in solving reasoning tasks such as math or coding in low-resource languages \citep{shi2023language,qin2023cross}.
This tendency primarily stems from the fact that LMs are predominantly trained on corpora comprised of a few high-resource languages~\citep{touvron2023llama1,touvron2023llama}. This results in low-resource languages being represented as long-tail knowledge~\citep{lazaridou2021mind, kandpal2023large}.

Prior works have mainly approached this problem by adapting English-centric LMs to other languages through continual training on the target language \citep{marchisio-etal-2023-mini,oba-etal-2023-second,zhu2023extrapolating, kew2023turning}. However, scaling this approach to a large number of languages is challenging, as it requires targeted training corpora for each language. This issue is particularly pronounced for LMs such as MetaMath \citep{yu2023metamath} and Orca 2 \citep{mitra2023orca}, which have undergone continuous domain-specific adaptation from Llama 2~\citep{touvron2023llama}. These specialized, domain-specific datasets are typically in English, complicating multilingual support for the underlying LM.

\begin{figure}[t!]
\includegraphics[width=1\linewidth]{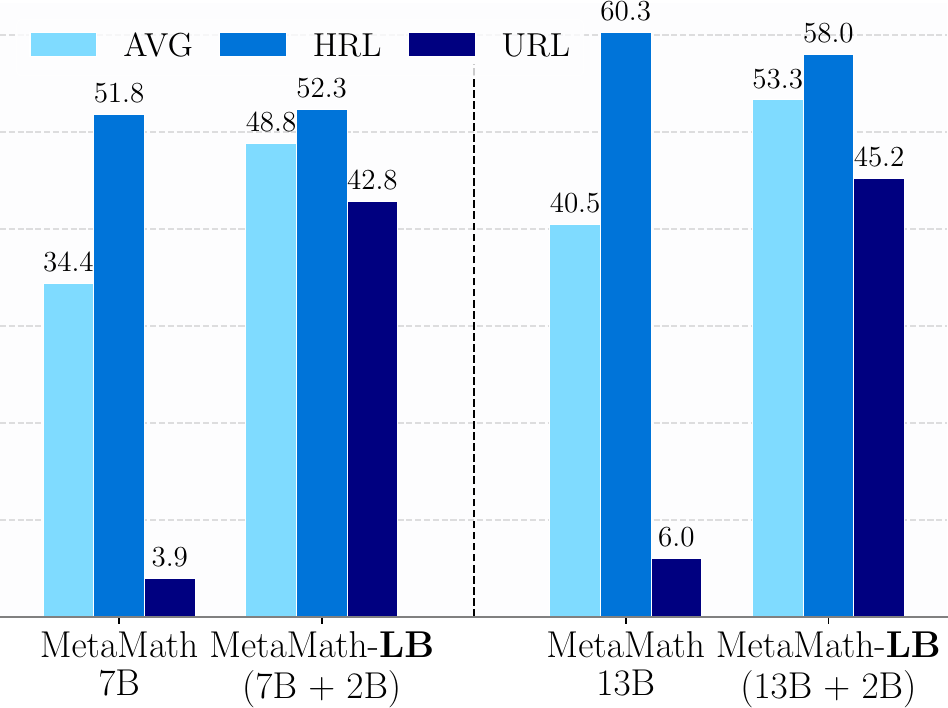}
\centering
\caption{MGSM accuracy (\%) of MetaMath models and models aligned with mT5-XL encoder (2B) via \method (\textbf{LB}). In addition to the average (\textsc{avg}) accuracy, we also report the average accuracy of high-resource languages (\textsc{hrl}) and underrepresented languages (\textsc{url}) classified by \citet{shi2023language}.}
\label{fig:main_metamath}
\vspace{-3mm}
\end{figure}

In this paper, we introduce \method, a novel approach that adapts LMs to solve multilingual reasoning tasks without \textit{explicitly} training on multilingual data. Inspired from the multimodal literature that integrates two independently pretrained modalities \citep{alayrac2022flamingo, li2023blip2, merullo2023linearly, liu2023llava, fuyu-8b}, we leverage the encoder from mT5 \citep{xue-etal-2021-mt5} and introduce a small number of trainable parameters between the encoder and the target LM. Most importantly, our approach does not require multilingual supervision and solely relies on English data while generalizing to multiple languages during test time, resembling zero-shot cross-lingual transfer \citep{pires-etal-2019-multilingual, conneau-etal-2020-unsupervised, xue-etal-2021-mt5}.

We demonstrate the effectiveness of \method by applying our method to LMs specialized in diverse reasoning tasks of mathematical reasoning, code completion, logical reasoning. Our empirical results show \method substantially enhances the multilingual reasoning performance of LMs. 
For example, \method applied to MetaMath-13B leveraging mT5-XL encoder (2.2B) boosts the average accuracy on MGSM \citep{shi2023language} from 40.5\% to 53.5\%, matching the performance of PaLM-540B \citep{chowdhery2023palm}, which stands at 51.3\%. We observe \method also significantly boosts LM performance on reasoning datasets that require intrinsic linguistic understanding such as specific subtasks of Big-Bench Hard \citep{suzgun-etal-2023-challenging} and XCOPA \citep{ponti-etal-2020-xcopa}.

We hypothesize that the effectiveness of \method is anchored in the language-agnostic characteristics of multilingual representations \citep{pires-etal-2019-multilingual, libovicky-etal-2020-language}. By mapping these representations to the target LM's input space, we conjecture that the LM is able to grasp the semantics of these representations. Since these representations are language-neutral, understanding them allows the LM to become less dependent on the specific language of the input, thereby enabling it to tackle tasks in languages it rarely encountered during pretraining. Our empirical analysis of \method, using principal component analysis (PCA) and qualitative methods, supports this hypothesis.

\begin{figure*}[ht!]
\includegraphics[width=1\textwidth]{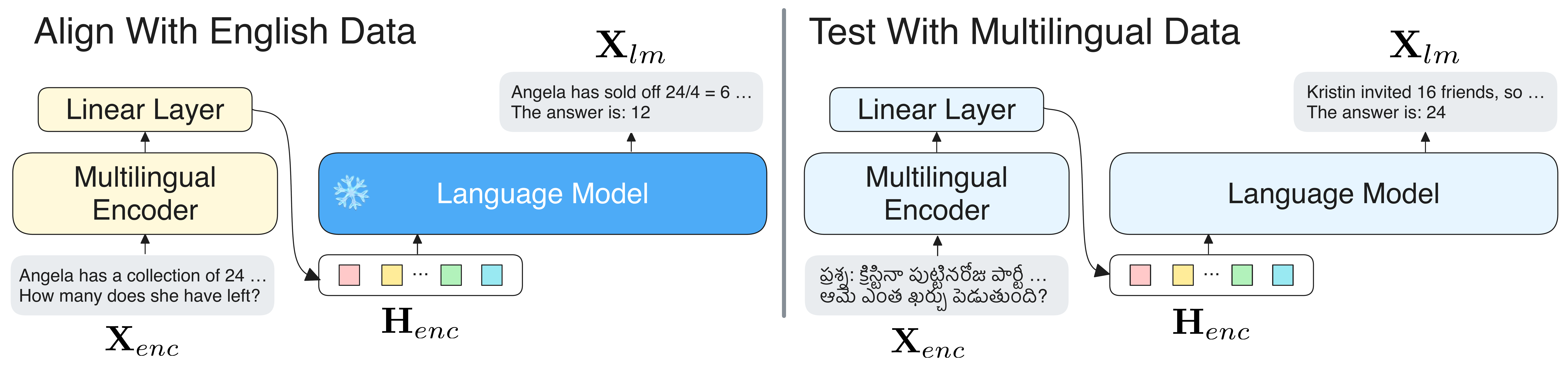}
\centering
\caption{Overview of \method. \textbf{Left}: A multilingual encoder with an added linear layer is aligned with the target language model using English data. We keep the language model frozen, whereas the linear layer is trainable. The multilingual encoder is trainable when adapting pretrained LMs and frozen when adapting finetuned LMs. \textbf{Right}: In test time, a \method model can effectively solve multilingual reasoning tasks.}
\label{fig:method}
\vspace{-3mm}
\end{figure*}

\section{Related Work}
\subsection{English-centric Language Models}

Previous works have enhanced the reasoning capabilities of LMs in mathematical reasoning, code generation, and logical reasoning \citep{mukherjee2023orca, azerbayev2023llemma, yu2023metamath, rozière2023code, mitra2023orca}. However, the majority of these LMs are derived from English-centric LMs \citep{touvron2023llama1, touvron2023llama} and adapted with English domain-specific datasets. Consequently, these LMs inherit limited proficiency in low-resource languages, which results in a significant performance discrepancy between high-resource and low-resource languages. The motivation of our proposed method, \method, is to narrow this gap.

\subsection{Zero-shot Cross-lingual Transfer}
Multilingual models demonstrate remarkable zero-shot cross-lingual transfer capability \citep{conneau-etal-2020-unsupervised, xue-etal-2021-mt5, fitzgerald-etal-2023-massive}. This ability significantly aids the low-resource language community, allowing multilingual models to handle tasks across a wide range of languages after being finetuned on high-resource languages. Our method, which leverages multilingual models, exhibits a similar ability, as it can enhance the reasoning capabilities of LMs across multiple languages while solely relying on English data during adaptation.

\subsection{Aligning Pretrained Representations}

The concept of combining independently pretrained representations has been widely explored in the realm of cross-modal alignment \citep{alayrac2022flamingo, li2023blip2, merullo2023linearly, liu2023llava, fuyu-8b}. These works focus on aligning visual encoder models with LMs to facilitate the visual comprehension abilities of LMs. In a concurrent work, \citet{bansal2024llm} align two large language models to augment each other's capabilities. One of their experiments showcases that aligning a multilingual LM with another LM can lead to improved performance in multilingual tasks. However, in contrast to our method, their approach employs multilingual supervision for aligning.

\section{\method} \label{method}
\subsection{Hypothesis}
Previous works show that representations of multilingual encoder models are moderately language-agnostic (or language-neutral) out-of-the-box, facilitating zero-shot cross-lingual transfer \citep{pires-etal-2019-multilingual, libovicky-etal-2020-language}. Based on this premise, we hypothesize that by aligning a multilingual encoder model to the space of an LM, the LM will be able to understand the semantics of the languages supported by the multilingual encoder \textit{without} training on a large set of languages.

To test this hypothesis, we align multilingual encoder models with LMs using only English corpora (\textbf{Left} of Figure \ref{fig:method}). Then, we evaluate the aligned models using non-English inputs to determine if the LMs exhibit enhanced capabilities in multilingual tasks (\textbf{Right} of Figure \ref{fig:method}).

\subsection{Model Architecture} \label{model_architecture}
Building on the findings of previous works, where effective cross-modal understanding has been achieved by mapping representations from non-linguistic modalities to the soft prompts \citep{lester-etal-2021-power} of LMs \citep{merullo2023linearly, liu2023llava, fuyu-8b}, \method maps the final hidden states of multilingual encoders to the soft prompts of LMs to attain multilingual understanding of the target LM. Following the studies above, we adopt the setting of using a single linear layer as the mapping function and append one trainable token to the end of the soft prompt as an \texttt{[EOS]} (end of sequence) token. Therefore, given the input tokens (padded if necessary) of the encoder $\mathbf{X}_{enc}$, the derived soft prompt $\mathbf{H}_{enc}$ is equivalent in sequence length to $\mathbf{X}_{enc} + 1$, and has the same dimensionality as the hidden state of the language model. Any tokens in $\mathbf{H}_{enc}$ that originate from padding tokens of $\mathbf{X}_{enc}$ are masked for the target LM. We ablate the effect of using more complex architectures in Appendix \ref{ablations}.

Language modeling objective of \method resembles that of the \say{prefix LM} explored by \citet{2020t5}, 
as the encoder input tokens $\mathbf{X}_{enc}$ can be interpreted as prefix tokens upon which the target tokens $\mathbf{X}_{lm}$ is conditioned. Formally, given the encoder input tokens $\mathbf{X}_{enc}$, the language modeling likelihood of the target tokens $\mathbf{X}_{lm}$ is denoted as:
\begin{equation}
    p(\mathbf{X}_{lm}|\mathbf{X}_{enc}) = \prod_i^Lp(x_i|\mathbf{X}_{enc},x_{<i})
\end{equation}
where $L$ is the sequence length of $\mathbf{X}_{lm}$, and $x_i$ is the $i$th token of $\mathbf{X}_{lm}$.

\section{Main Experiments} \label{main_experiments}
\subsection{Overview} \label{experiments_overview}

We select four task categories for our experiments: (1) \textbf{mathematical reasoning}, (2) \textbf{code completion}, (3) \textbf{logical reasoning}, and (4) \textbf{commonsense reasoning}.\footnote{Experiments on commonsense reasoning can be found in Appendix~\ref{commonsense}.} For each task category, we apply \method to LMs specialized in reasoning derived from Llama~2~\citep{touvron2023llama}, such as MetaMath or Orca 2.\footnote{We test \method on general-domain LMs (Llama 2 and Mistral 7B \citep{jiang2023mistral}) in Appendix \ref{general_lms}.} We evaluate the models on existing multilingual benchmarks (e.g., MGSM) or translated English benchmarks (e.g., translations of HumanEval \citep{chen2021codex}). As the evaluation tasks necessitate both multilingual understanding abilities and advanced reasoning capabilities, this complexity poses a significant challenge for general multilingual LMs and English-centric LMs specialized in reasoning. On the contrary, models aligned with \method could take advantage of both.

Since \citet{touvron2023llama} disclose the language distribution of the pretraining data of Llama~2, this enables us to identify which languages are underrepresented in LMs initialized from Llama 2 weights. Throughout the paper, we classify a language as underrepresented if it comprises less than 0.1\% of Llama 2's pretraining data.

In all of our experiments, we use the encoders of mT5 \citep{xue-etal-2021-mt5} as the multilingual encoders due to their availability across a wide range of parameters,\footnote{270M (Small), 470M (Base), 820M (Large), 2.2B (XL) and 6.7B (XXL).} and their support for longer input sequences compared to other multilingual encoder models.\footnote{mT5 was trained on input size of 1024 tokens, but can take longer sequences due to its use of relative position embeddings \citep{shaw-etal-2018-self}.} Specifically, we use the \say{LM adaptated} checkpoints from \citet{vu-etal-2022-overcoming}. We align target LMs of 7B parameters with mT5-XL encoder to adapt 9B-sized models and 13B-sized LMs with mT5-XL encoder and mT5-XXL encoder to obtain 15B and 20B models, respectively. As \method adds a considerable amount of additional parameters to the target LM, we conduct an inference throughput analysis in Appendix \ref{throughput}. We also ablate the effect of the encoder parameter size and encoder model type in Appendix \ref{ablations}.

We use the original continual training data of the LM for \method when accessible (e.g., MetaMathQA for MetaMath). If unavailable, we opt for the closest publicly accessible dataset (e.g., OpenOrca \citep{OpenOrca} for Orca 2). In all our experiments, we fix the size of the training dataset to 200,000 instances. However, our ablation study on the effect of the training dataset size in Appendix \ref{ablations} suggests ~\method in practice may require much less data. We maintain the language model frozen throughout the alignment process for efficiency. We also freeze the encoder (except the embedding layer) for aligning finetuned LMs, whereas for pretrained LMs, we keep the encoder trainable.\footnote{We define \textit{finetuned} LMs as LMs trained on labeled corpora and \textit{pretrained} LMs as LMs trained on unlabeled corpora.} In Appendix \ref{ablations}, we provide further explanations for these choices and ablate the effects of freezing. We align the models by training on the prefix LM objective described in Section \ref{model_architecture}. In our preliminary experiments, we find that training on various lengths of $\mathbf{X}_{enc}$ is necessary to ensure robustness on inference time, as the language model is exposed to diverse lengths of $\mathbf{H}_{enc}$. 

We use a maximum input length ($\mathbf{X}_{enc}$) of 1,024 and a maximum target length ($\mathbf{X}_{lm}$) of 128 for training. For unlabeled data, we randomly vary the input length within the 1,024 window to introduce the LM to various lengths of $\mathbf{H}_{enc}$. For labeled data, the data naturally comes in diverse input lengths. On a machine equipped with four A100 80GB GPUs, training a 9B model takes less than four hours when the encoder layers are frozen, and under five hours when the entire encoder is trainable. In our main experiments, the setting where the encoder is fully trainable in the 20B models results in the maximum training time, which is approximately ten hours. Further training details are available in Appendix \ref{experiemental_details}.


\begin{table*}[!t]
\small
\centering
\begin{tabularx}{1\textwidth}{l|>{\centering\arraybackslash}X>{\centering\arraybackslash}X>{\centering\arraybackslash}X|>{\centering\arraybackslash}X>{\centering\arraybackslash}X>{\centering\arraybackslash}X>{\centering\arraybackslash}X>{\centering\arraybackslash}X>{\centering\arraybackslash}X>{\centering\arraybackslash}X|>{\centering\arraybackslash}X>{\centering\arraybackslash}X>{\centering\arraybackslash}X>{\centering\arraybackslash}X}
\toprule
 & \textbf{\textsc{avg}} & \textbf{\textsc{hrl}} & \textbf{\textsc{url}} &\textbf{\textsc{en}} & \textbf{\textsc{de}} & \textbf{\textsc{fr}} & \textbf{\textsc{es}} & \textbf{\textsc{ru}} & \textbf{\textsc{zh}} & \textbf{\textsc{ja}} & \textbf{\textsc{th}} & \textbf{\textsc{sw}} & \textbf{\textsc{bn}} & \textbf{\textsc{te}}\\
\midrule
Lang. Freq. (Llama 2, \%) & - & - & - & 89.7 & 0.17 & 0.16 & 0.13 & 0.13 & 0.13 & 0.10 & \multicolumn{4}{c}{\textsc{Less Than 0.005}}\\
\midrule
\midrule
\multicolumn{15}{c}{\textsc{Few-Shot Cross-lingual CoT}}  \\
\midrule
Llama 2-7B & 9.1 & 12.1 & 3.9 & 15.2 & 11.6 & 13.2 & 11.2 & 11.6 & 11.2 & 10.8 & 7.2 & 5.2 & 3.2 & 0.0\\
XGLM-7.5B & 1.5 & 1.6 & 1.2 & 0.4 & 1.6 & 1.2 & 1.6 & 2.0 & 2.8 & 1.6 & 2.0 & 0.4 & 1.2 & 1.2 \\
mT5-XXL (13B)  &2.9 & 3.5& 2.0 & 3.6 & 2.4 & 4.0 & 3.6 & 2.8 & 3.6 & 4.4 & 2.8 & 1.2 & 3.2 & 0.8\\
BLOOM-7.1B & 2.4& 2.6& 2.0 & 3.6 & 1.2 & 3.6 & 2.4 & 2.0 & 3.2 & 2.0 & 0.0 & 2.4 & 2.8 & 2.8 \\
BLOOM-7.1B-PP2 &2.3& 2.5& 1.9 & 4.8 & 1.2 & 2.0 & 2.0 & 1.6 & 4.0 & 1.6 & 0.8 & 2.8 & 2.0 & 2.0 \\
PaLM-540B & 51.3 & 52.3 & 46.8 & 62.4 & 53.6 & 51.2 & 58.0 & 55.6 & 46.0 & 49.6 & 49.6 & 44.4 & 46.4 & 46.8\\
\midrule
Llemma-7B & \textbf{21.6}& \textbf{29.9} & 7.2 & \textbf{44.8} & \textbf{27.2} & \textbf{33.2} & \textbf{29.2} & \textbf{26.0} & \textbf{26.4} & \textbf{22.4} & 14.0 & 8.4 & 6.4 & 0.0 \\
Llemma-\methodtable-9B & 20.4 & 22.5& \textbf{16.7} & 34.8 & 23.6 & 26.8 & 22.4 & 18.8 & 16.0 & 15.2 & \textbf{20.8} & \textbf{17.6} & \textbf{12.4} & \textbf{16.0}\\ 
\midrule
Llemma-34B & 35.6 & 46.3 & 16.7 & 58.0 & 48.0 & 46.8 & 48.0 & 47.2 & 36.8 & 39.6 & 28.4 & 27.2 & 11.2 & 0.0 \\
\midrule
\midrule
\multicolumn{15}{c}{\textsc{Zero-shot CoT}}  \\
\midrule
MathOctopus-7B & 37.1& 42.7 & 27.2 & 51.6 & 40.0 & 38.4 & 47.2 & 42.4 & 44.0 & 35.6 & 39.2 & 31.6 & 37.2 & 0.8 \\
MathOctopus-13B & 42.9 & 48.6 & 32.9 & 50.8 & 49.2 & 50.4 & 52.8 & 47.2 & 52.4 & 37.2 & 44.4 & 40.4 & 46.4 & 0.4 \\
BLOOM-7.1B-MM & 16.7& 21.7 & 7.8 & 41.2 & 19.6 & 24.4 & 26.8 & 9.6 & 21.2 & 9.2 & 0.8 & 15.6 & 6.8 & 8.0 \\
\midrule
MetaMath-7B & 34.4& 51.8 & 3.9 & \textbf{64.8} & \textbf{57.6} & \textbf{55.6} & 56.4 & 50.4 & 42.4 & 35.6 & 4.0 & 6.4 & 4.4 & 0.8 \\
MetaMath-\methodtable-9B & \textbf{48.8}& \textbf{52.3}& \textbf{42.8} & 63.2 & 50.8 & 52.4 & \textbf{58.0} & \textbf{56.4} & \textbf{45.2} & \textbf{40.0} & \textbf{50.4} & \textbf{43.2} & \textbf{42.8} & \textbf{34.8} \\
\midrule
MetaMath-13B & 40.5& \textbf{60.3} & 6.0 & \textbf{70.4} & \textbf{64.4} & \textbf{65.2} & \textbf{63.6} & \textbf{60.0} & 50.8 & \textbf{47.6} & 4.8 & 11.6 & 6.8 & 0.8 \\
MetaMath-\methodtable-15B & 53.5 & 58.0 & 45.2 &67.6 & 63.6 & 61.6 & 63.2 & \textbf{60.0} & 48.0 & 42.0 & 52.8 & 41.6 & 50.0 & 36.4 \\
MetaMath-\methodtable-20B & \textbf{55.8} & 58.7& \textbf{50.7} & 66.4 & 64.0 & 64.0 & 60.4 & 58.8 & \textbf{52.4} & 45.2 & \textbf{53.6} & \textbf{49.2} & \textbf{52.8} & \textbf{47.2} \\
\bottomrule
\end{tabularx}
\caption{
Accuracy (\%) on MGSM. Alongside average (\textsc{avg}) accuracy, we also report average accuracy of high-resource languages (\textsc{hrl}) and underrepresented languages (\textsc{url}) classified by \citet{shi2023language}. We include the language distribution of Llama 2 for reference. For pretrained models (\textbf{Top}), we prompt with 8-shot cross-lingual chain-of-thought (CoT) reasoning exemplars, except for PaLM-540B, for which we reference the 6-shot cross-lingual CoT performance reported by \citet{shi2023language}. For finetuned models (\textbf{Bottom}), we evaluate zero-shot. The PP2 and MM suffixes denote models trained on Proof-Pile-2 and MetaMath, respectively. We compare \method models (\textbf{LB}) to their original LMs and highlight the best-performing numbers in \textbf{bold}.
}
\label{table:mgsm_main}
\end{table*}

\subsection{Mathematical Reasoning} 
\subsubsection{Experimental Setup}
\paragraph{Evaluation Datasets}
\textbf{MGSM} \citep{shi2023language} comprises grade school math word problems in 11 typologically diverse languages, human translated from a sample of GSM8K \citep{cobbe2021gsm8k}. For evaluating pretrained LMs, we adopt the cross-lingual transfer chain-of-thought (CoT) reasoning \citep{wei2022chain} setting (\textsc{native-exemplars} + \textsc{en-cot}) from \citet{shi2023language}, where the few-shot exemplars are given in the target language, but the CoT rationales to solve the exemplars are provided in English. For finetuned LMs, we evaluate in zero-shot\footnote{Here, the term \textit{zero-shot} refers to the lack of few-shot examples.} setting. Additional evaluation on MSVAMP \citep{chen2023breaking} is available in Appendix \ref{additional_evaluations}.

\paragraph{Language Models} \textbf{Llemma} \citep{azerbayev2023llemma} is a set of LMs for mathematics, continually pretrained from Code Llama \citep{rozière2023code} on Proof-Pile-2, a mixture of scientific papers, web data containing mathematics, and mathematical code. \textbf{MetaMath} \citep{yu2023metamath} was finetuned from Llama 2 \citep{touvron2023llama} on MetaMathQA, a mathematical dataset based on GSM8K and MATH \citep{hendrycksmath2021}. As both Proof-Pile-2 and MetaMathQA are publicly available, we apply \method using samples of their respective training datasets.

\paragraph{Baselines} \textbf{Llama 2} \citep{touvron2023llama} is an English-centric LM in which 89.7\% of the pretraining data consists of English but has shown considerable performance on non-English languages \citep{lai2023okapi}. \textbf{mT5}\footnote{We use the language model checkpoint from \citet{vu-etal-2022-overcoming}.} \citep{xue-etal-2021-mt5}, \textbf{XGLM} \citep{lin2022fewshot}, and \textbf{BLOOM} \citep{workshop2022bloom} are massively multilingual LMs. \textbf{MathOctopus} \citep{chen2023breaking} is an LM for multilingual mathematical reasoning. It was initialized from Llama 2 and finetuned on translations of the GSM8K dataset across ten languages. The ten languages seen by MathOctopus overlap with the 11 languages included in MGSM, except Telugu. We use their best-performing checkpoints, xRFT-MathOctopus$^P$, which were further enhanced by data augmentation through rejection sampling \citep{yuan2023scaling}. We also report the performance of BLOOM models further trained on the training sets of \method models: \textbf{BLOOM-Proof-Pile-2 (PP2)} and \textbf{BLOOM-MetaMath (MM)}. This is done to confirm that the capabilities of \method models are derived from the LMs' inherent strength rather than solely from the training set utilized by \method. Detailed hyperparameters for training the BLOOM models are available in Appendix \ref{experiemental_details}. We additionally report the performance of \textbf{PaLM} \citep{chowdhery2023palm} measured by \citet{shi2023language}. Similar to Llama 2, PaLM was pretrained on English-heavy corpora.

\subsubsection{Results}
Table \ref{table:mgsm_main} shows the evaluation results of baselines and \method models on MGSM. We highlight six main observations. (1)~Llama 2, Llemma, and MetaMath exhibit critical performance degradation across languages that are underrepresented in the training data of Llama~2. (2) Despite this, massively multilingual LMs underperform Llama~2, even in the context of underrepresented languages.\footnote{Note BLOOM models were not trained in German, Russian, Japanese, and Thai.} This disparity underscores the robust mathematical reasoning capabilities inherent in Llama 2 and absent in BLOOM, XGLM, and mT5. (3) \method enhances the multilingual performance of Llemma and MetaMath, especially in underrepresented languages. Most notably, \method is able to bring Llemma and MetaMath performance on Telugu (\textsc{te}) from zero or near zero to a range comparable to other languages. (4) \method may degrade performance on high-resource languages, with Llemma-\methodtable-9B's English (\textsc{en}) performance drop being particularly pronounced. We provide our speculations on the cause of this phenomenon in Section \ref{degradation}.
(5) Mathematical reasoning capabilities of \method models come from their original LMs, not their training data. This is evident from BLOOM-7.1B-PP2 and BLOOM-7.1B-MM underperforming Llemma-\methodtable-9B and MetaMath-\methodtable-9B, respectively, by a large margin. (6) Surprisingly, despite only being trained on English math data, our MetaMath-\methodtable models are competitive against MathOctopus models, which were finetuned on translations of GSM8K on ten out of 11 languages supported by MGSM. Most importantly, performance of MathOctopus models drop to near zero on Telugu (\textsc{te}), an unseen language by MathOctopus. On the other hand, \method models show robust performance on all 11 languages, suggesting that even without multilingual supervision \method generalizes to the large scale of languages included in the multilingual pretraining of the encoders.


Overall, \method models demonstrate outstanding performance against baselines. \method models vastly outperform similar-sized multilingual models, establishing \method as a viable approach for developing mathematical reasoning models for low-resource languages.
We provide an example of a CoT rationale generated by MetaMath-\methodtable in Appendix \ref{cot}.

\begin{table*}[ht!]
\small
\centering
\begin{tabularx}{1.0\textwidth}{l|>{\centering\arraybackslash}X|>{\centering\arraybackslash}X>{\centering\arraybackslash}X>{\centering\arraybackslash}X>{\centering\arraybackslash}X>{\centering\arraybackslash}X>{\centering\arraybackslash}X}
\toprule

 & \textbf{AVG} & \textbf{EN} & \textbf{SW} & \textbf{BN} & \textbf{PA}  & \textbf{TE}  & \textbf{UR} \\
\midrule
Llama 2-7B & 4.6 & 9.8 & 3.7 & 3.0 & 3.0 & 3.0 & 4.9 \\
BLOOM-7.1B & 4.9 & 7.3 & 2.4 & 5.5 & 4.3 & 6.1 & 3.7 \\
BLOOM-7.1B-StarCoder & 4.5 & 6.7 & 3.0 & 4.3 & 4.3 & 3.7 & 4.9 \\
\midrule
Code Llama-7B & 16.9 & \textbf{36.0} & 14.0 & 14.6 & 10.4 & 9.8 & 16.5\\
Code Llama-\methodtable-9B & \textbf{19.4} & 31.7 & \textbf{17.1} & \textbf{15.2} & \textbf{18.3} & \textbf{15.2} & \textbf{18.9} \\
\midrule
Code Llama-13B & 19.6 & 40.2 & 15.2 & 17.7 & 12.2 & 12.8 & 19.5 \\
Code Llama-\methodtable-15B & \textbf{23.6} & \textbf{41.5} & 20.1 & \textbf{20.1} & 19.5 & \textbf{19.5} & \textbf{20.7} \\
Code Llama-\methodtable-20B & 23.5 & 36.0 & \textbf{24.4} & \textbf{20.1} & \textbf{20.7} & 18.9 & \textbf{20.7} \\
\midrule
Code Llama-34B & 22.7 & 43.9 & 17.1 & 23.8 & 15.9 & 12.8 & 22.6 \\
\bottomrule
\end{tabularx}
\caption{
Pass@1 scores on \textit{anonymized} versions of HumanEval and HumanEval-MT. The models were evaluated on zero-shot code completion using greedy decoding. We compare \method (\methodtable) models to their original checkpoints and highlight the best-performing numbers in \textbf{bold}.
}
\label{table:humaneval_main_no_anon}
\end{table*}

\subsection{Code Completion} \label{code completion}

\subsubsection{Experimental Setup}
\paragraph{Evaluation Datasets}
Leveraging the strong translation capabilities of GPT-4 \citep{openai2023gpt4}, as demonstrated in the study by \citet{jiao2023chatgpt}, we extend \textbf{HumanEval} \citep{chen2021codex}, a set of hand-written programming problems, into five underrepresented languages: Swahili, Bengali, Punjabi, Telugu, and Urdu. We name the resulting dataset \textbf{HumanEval-MT}. We select the five languages among those with reported MMLU \citep{hendryckstest2021} performance in the technical report of GPT-4 while also being included in the pretraining corpora of BLOOM. This choice is made to acquire high-quality translations from GPT-4, and provide BLOOM-based baselines a level playing field. To guide GPT-4 to only translate the embedded natural language instruction of the docstring while not modifying the code segments, we prompt GPT-4 with two human-annotated examples.\footnote{We provide entire functions as inputs to GPT-4 to give it as much context as possible for accurate translation.} Subsequently, the generated translations are executed in a Python interpreter environment to assert the absence of syntax errors. We provide translation quality estimation of HumanEval-MT in Appendix \ref{qe_humaneval-MT}.

The examples within the HumanEval dataset often feature self-explanatory function names. This suggests that language models could potentially complete the associated code segments successfully, even without accurately comprehending the natural language comments embedded within them. Therefore, we evaluate the models on \textit{anonymized} versions of each language set, wherein the target function names of the code segments are uniformly altered to \say{\texttt{func}}. An example of the anonymization and evaluation result on the original version is available in Appendix \ref{additional_evaluations}.

\paragraph{Language Models} \textbf{Code Llama} \citep{rozière2023code} is a family of models initialized from Llama~2 model weights and pretrained on a code-heavy dataset. In our experiments, we use Code Llama-Python models, which were further pretrained on a Python-heavy dataset. Since the datasets used to pretrain the Code Llama models are not publicly available, we sample from the Python subset of StarCoder data \citep{li2023starcoder} as the training dataset for \method.

\paragraph{Baselines} We use \textbf{Llama 2} and \textbf{BLOOM} models as the baselines. Both models contain code data within their pretraining corpora. Additionally, we report the performance of \textbf{BLOOM-StarCoder}, a BLOOM model continually pretrained on the sample of StarCoder data used to train \method models.

\subsubsection{Results}
Table \ref{table:humaneval_main_no_anon} presents the Pass@1 scores on HumanEval and HumanEval-MT.
Code Llama-\methodtable models show consistent improvements over Code Llama across all underrepresented languages. Moreover, \method models can match their larger baselines in terms of average scores: the 9B model slightly underperforms Code Llama-13B by 0.2\%, while both the 15B and 20B models surpass Code Llama-34B. BLOOM trained on StarCoder data did not demonstrate noticeable improvements, re-emphasizing that the strengths of \method models predominantly stem from the capabilities of original LMs.

\subsection{Logical Reasoning} 

\subsubsection{Experimental Setup}
\paragraph{Evaluation Datasets}

We assess logical reasoning capabilities with \textbf{Big-Bench Hard (BBH)} \citep{suzgun-etal-2023-challenging} and \textbf{Big-Bench Hard Bengali (BBH-BN)} \citep{shafayat2024benqa}. BBH is a collection of challenging subtasks where the application of chain-of-thought (CoT) reasoning has the potential to enhance performance substantially. BBH-BN translates 14 of the 23 subtasks of BBH into Bengali. To facilitate meaningful comparison, we evaluate only on the 14 subtasks supported by BBH-BN for BBH.\footnote{List of the 14 subtasks is available in Appendix \ref{additional_evaluations}.}

\paragraph{Language Models}
\textbf{Orca 2} \citep{mitra2023orca} was finetuned on Llama 2 with a collection of datasets augmented with reasoning traces of GPT-4 as well as fully synthetic datasets created with GPT-4. Orca 2 effectively improved the reasoning abilities of smaller LMs on complex tasks demanding advanced reasoning in zero-shot settings. As the training dataset of Orca 2 is not publicly available, we sample the training data for \method from the OpenOrca dataset \citep{OpenOrca}. OpenOrca follows the data distribution of the first iteration of Orca \citep{mukherjee2023orca}. We employ CLD3\footnote{\href{https://github.com/google/cld3}{github.com/google/cld3}} to filter any non-English data that mainly derives from translation datasets to ensure the \textit{zero-shot} setting of our experiments. Examples were included if their input text had a 99\% or greater probability of being English, while their target text also had a 95\% or greater chance of being English. A slightly lower threshold was adopted for the target text to not falsely filter single-word responses, which CLD3 exhibits lower confidence.

\paragraph{Baselines}
In our evaluation of BBH, we assess whether Orca 2-\methodtable models could acquire multilingual comprehension while retaining the zero-shot CoT capabilities of Orca 2. However, from our limited testing, we found that no existing open multilingual LMs could generate CoT reliably in a zero-shot setting. Consequently, they were not included as baselines. We do report the performance of \textbf{BLOOM-OpenOrca}, a BLOOM model further trained on the same training set as Orca 2-\methodtable.

\begin{table}[t!]
\centering
\begin{tabularx}{\linewidth}{l|>{\centering\arraybackslash}X>{\centering\arraybackslash}X}
\toprule
 & \textbf{EN} & \textbf{BN} \\
\midrule
BLOOM-7B-OpenOrca & 35.8 & 31.2 \\
\midrule
Orca 2-7B & \textbf{53.9} & 36.7 \\
Orca 2-\methodtable-9B & 46.9 & \textbf{41.8} \\
\midrule
Orca 2-13B & \textbf{57.9} & 41.7 \\
Orca 2-\methodtable-15B & {55.2} & \textbf{45.4} \\
Orca 2-\methodtable-20B & {53.1} & \textbf{45.4} \\

\bottomrule
\end{tabularx}
\caption{
Accuracy (\%) on BBH (English) and BBH-BN (Bengali). We report the average accuracy across 14 subtasks. We compare \method (\methodtable) models to their original checkpoints and highlight the best-performing numbers in \textbf{bold}.
}
\label{table:bbh_main}
\end{table}

\subsubsection{Results}
Table \ref{table:bbh_main} shows the average accuracy across the subtasks for BBH and BBH-BN. Orca 2-\methodtable-9B model shows considerable improvement in BBH-BN, surpassing the larger Orca 2-13B model. We report the performances on each subtask in Table \ref{table:bbh1} and Table \ref{table:bbh2}. Remarkably, \method significantly increases the performance on BBH-BN subtasks that require intrinsic linguistic understanding such as \textsc{Causal Judgement} and \textsc{Snarks}. \textsc{Causal Judgement} requires the model to comprehend a short story and determine how a typical human would answer a given question. For \textsc{Snarks}, the model is given two nearly-identical sentences and is required to indentify which one contains sarcasm. This observation, together with the results from the evaluation on XCOPA \citep{ponti-etal-2020-xcopa}, a commonsense reasoning dataset, suggests that \method models are robust in grasping and interpreting nuanced linguistic details. Evaluation results on XCOPA are provided in Appendix \ref{additional_evaluations}. Figure \ref{fig:cot_causal} shows an example of a \method model correctly solving \textsc{Causal Judgement}.

\section{Analysis of \method}

Based on the empirical evidence presented in the previous sections, we assert that \method effectively enhances LMs' capability to address multilingual tasks without multilingual training, especially for low-resource languages. This section presents two fundamental observations that further reinforce the hypothesis outlined in Section~\ref{method}.  

\subsection{PCA}

The hypothesis primarily attributes the feasibility of \method to the sufficiently language-agnostic representations of multilingual encoders. If the conjecture holds, given a \method model, the LM's output representation of the soft prompt $\mathbf{H}_{enc}$ should also exhibit language-agnostic characteristics. It stands to reason that the LM would not arbitrarily introduce additional language-specific features to a language-neutral input derived from the multilingual encoder.

\begin{figure}[t!]
\includegraphics[width=1\linewidth]{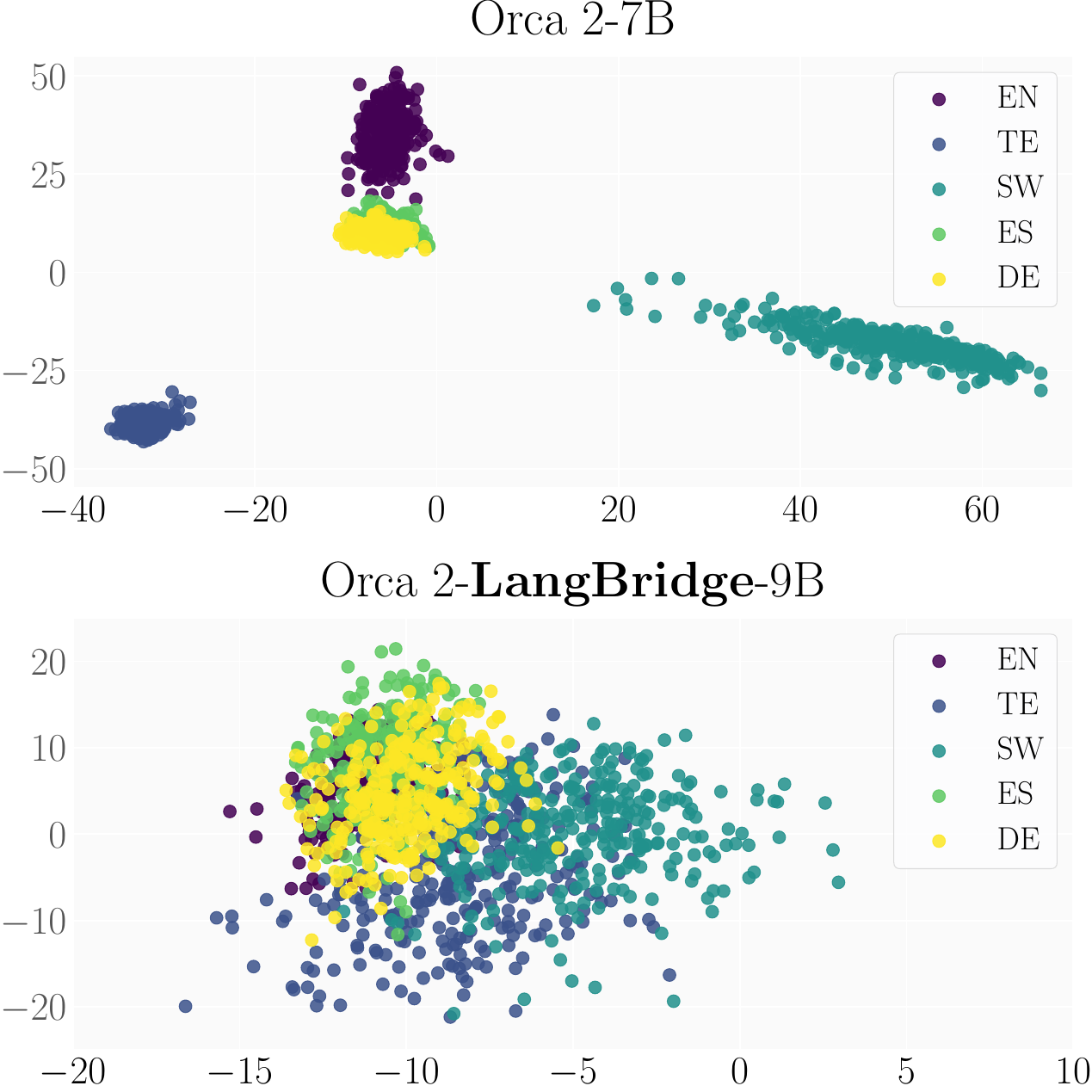}
\centering
\caption{First two principal components of pooled output representations obtained with 300 FLORES samples per language. Note that the scales of the two subplots differ.}
\label{fig:pca}
\vspace{-3mm}
\end{figure}

Therefore, to verify if the \method models truly have language-agnostic output representations, we apply PCA to the mean pooled output representations from a \method model and compare them with those from the original English-centric LM. Figure \ref{fig:pca} shows the first two principal components of pooled output representations obtained with FLORES \citep{goyal-etal-2022-flores, nllbteam2022language}, a parallel corpus. For Orca 2, high-resource languages, English (\textsc{en}), Spanish (\textsc{es}), and German (\textsc{de}), are mapped closely together. Underrepresented languages, Telugu (\textsc{te}) and Swahili (\textsc{sw}), exhibit a more distant mapping in the representation space, forming three clusters.\footnote{Note that Flores Swahili is in Latin script, the same as the three high-resource languages.} Conversely, for Orca 2-\method, all languages are mapped into a single cluster, indicating that the representations of $\mathbf{H}_{enc}$ maintain a relatively language-neutral status.

\begin{figure}[t!]
\includegraphics[width=1\linewidth]{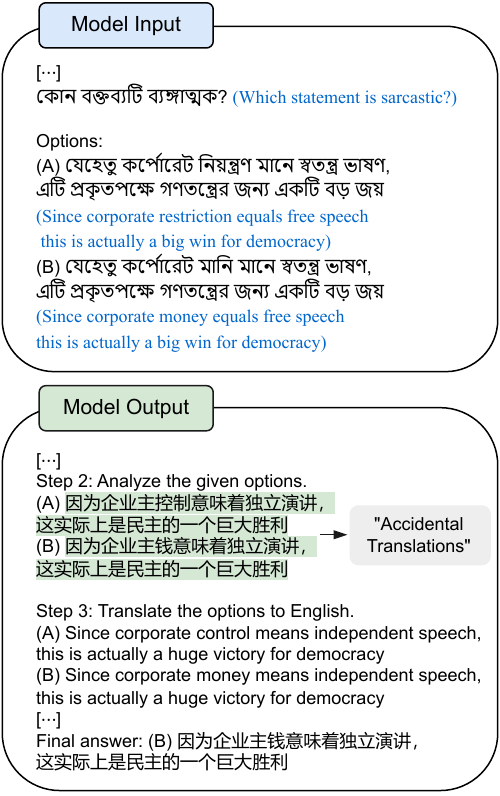}
\centering
\caption{Example of accidental translation of an Orca 2-\method model prompted with the \textsc{Snark} subset of BBH-BN. Portions of the input prompt and several rational steps in the output are truncated for brevity. Translations are provided in \blue{(blue with parenthesis)} wherever required.}
\label{fig:accidental}
\vspace{-3mm}
\end{figure}

\subsection{Accidental Translations}

Figure \ref{fig:accidental} illustrates an example of \say{accidental translation} \citep{xue-etal-2021-mt5} by the Orca 2-\method-15B model. Despite being given the options in Bengali, the \method model perceives the options as Chinese and recites them in Chinese. With Bengali as input, accidental translation in a third language other than English or Bengali suggests that multiple languages may have similar representations in $\mathbf{H}_{enc}$ \citep{li-murray-2023-zero}. Nonetheless, outputs in languages other than English are uncommon for Orca 2-\method models.
We conduct a qualitative analysis on the CoTs generated by Orca 2-\method-15B for BBH-BN \textsc{Snarks} subtask, which we find is the combination with the most frequent accidental translations. Out of 178 CoT rationales generated, only eight examples contained accidental translations in Chinese, Danish, Hindi, Japanese, Marathi, and Turkish, with one or two examples per language. Additionally, seven examples had Bengali in their CoT rationales. The relatively high frequency of Bengali indicates that $\mathbf{H}_{enc}$ does not exhibit a perfectly language-agnostic behavior, and for some examples, the LM could discern the input language as Bengali. This is not ideal, as it suggests that the LM had to comprehend the input in Bengali, a language in which the LM lacks proficiency. We believe that \method performance can be further enhanced by relieving the \textit{zero-shot} constraint and adapting the mT5 encoder to have enhanced language-neutrality \citep{reimers-gurevych-2020-making,feng-etal-2022-language} prior to alignment with the LM. However, we leave this exploration for future study.

\section{Discussion}
\subsection{HRL Performance Degradation} \label{degradation}
In some cases, \method models exhibit performance degradation in high-resource languages compared to their target LMs. While we leave further investigation for future studies, we outline some conjectures regarding potential causes for these performance declines. 
\begin{itemize}
    \item Given that the target LM is already proficient in a certain language, the text representations derived from the soft prompts may be more challenging for the LM to interpret compared to those from the LM's native embedding layer.
    \item Due to limited computing budget, \method models were trained with maximum input sequence length of 1024 ($\mathbf{X}_{enc}$) and maximum output sequence length of 128 ($\mathbf{X}_{lm}$). This is considerably shorter than what the original language models were trained on, especially for language models trained solely on unlabeled data such as Llemma.
    \item We fixed the training hyperparameters for all of the \method models regardless of the target LM for the sake of consistency and to save computing budget. This could have been suboptimal for some \method models.
    \item For Code Llama and Orca 2, their original training corpora are inaccessible. Therefore, we opt for the closest alternative as stated in Section \ref{experiments_overview}. Nonetheless, a minor distribution shift during LM training and \method training is inevitable.
\end{itemize}

\subsection{Multilingual CoT Capabilities?}
As \method solely utilizes English data for training, it is anticipated that the intermediate reasoning steps, or CoT, of \method models would predominantly be in English. Given that \method targets English-centric LMs specialized in reasoning, we conjecture that English CoT is one of the key components that contribute to the competitive performance of \method models. This claim is supported by the findings of \citet{shi2023language}, where they observe English CoT consistently leads to competitive results, and suggests that English CoT serves as a "useful baseline for future multilingual reasoning work".

While we believe incorporating non-English data into the \method training data could induce multilingual CoT capabilities in \method models, this approach does not align well with the original motivation of this work, which is to avoid the need for collecting multilingual reasoning data.

\section{Conclusion}

In this paper, we propose \method, a simple yet effective method of extending the capabilities of LMs to solve multilingual reasoning tasks without using multilingual supervision. We show that \method is surprisingly effective for enhancing multilingual reasoning capabilities for low-resource languages. Additionally, we provide analysis that indicates the effectiveness of \method is due to the language-agnostic nature of the multilingual representations. We hope our findings benefit the low-resource language users and spur further research advancing the development of LMs inclusive of the entire global community.

\section*{Limitations}
As \method solely utilizes English data for training, \method models may not be proficient in generating text in languages other than English. Although \method successfully narrows the performance gap between high-resource and low-resource languages, a noticeable performance gap remains. Also, while multilingual representations are known to have language-agnostic characteristics to some degree, our analysis and previous works suggest that there is room for improvements \citep{libovicky-etal-2020-language, feng-etal-2022-language}. While \method has the potential to generalize to all languages supported by the multilingual encoder, the extent to which \method enhances the reasoning capability of a specific language depends on two key factors: (1) The original proficiency of the LM in that particular language. (2) The proficiency of the encoder model in that particular language.

\section*{Ethical Considerations}
While we share \method models for open access, their terms for use or license adhere to those of the original LMs. The training datasets utilized in our research is primarily sourced from academic materials. As a result, we assess that the datasets likely contain a relatively fewer examples featuring offensive or personal information. Nevertheless, it is important to acknowledge that such content may be still present within the datasets.

\section*{Acknowledgements}
This work was partly supported by KAIST-NAVER Hypercreative AI Center (60\%) and Institute of Information \& communications Technology Planning \& Evaluation (IITP) grant funded by the Korea government (MSIT) (No.2022-0-00113, Developing a Sustainable Collaborative Multi-modal Lifelong Learning Framework, 20\%; No.2021-0-02068, Artificial Intelligence Innovation Hub, 20\%).

\bibliography{anthology,custom}

\appendix

\clearpage

\section{Additional Evaluation Results} \label{additional_evaluations}

\subsection{XCOPA} \label{commonsense}

Table \ref{table:copa_main} shows the evaluation results of Orca 2 and Orca 2-\method models on \textbf{COPA} \citep{roemmele2011choice} and \textbf{XCOPA} \citep{ponti-etal-2020-xcopa}, commonsense reasoning datasets. COPA is available in English, while XCOPA extends COPA to 11 languages. We do not include other LMs as baselines, as COPA was included in the training set of Orca 2, making it challenging to draw meaningful comparisons.

Despite reaching near-perfect accuracy for COPA, Orca 2 models' performance drops close to random chance (50\%) on some of the underrepresented languages of XCOPA. \method successfully decreases this performance degradation, except for Quechua (\textsc{qu}). The discrepancy is likely due to Quechua not being included in the 101 languages covered by mT5. This observation reemphasizes that the large-scale linguistic proficiency of \method models is primarily derived from mT5.  

\subsection{MSVAMP} \label{msvamp}
\textbf{MSVAMP} \citep{chen2023breaking} is a multilingual grade school level math word problem dataset translated from SVAMP \citep{patel-etal-2021-nlp} to 10 languages. We only evaluate MSVAMP in a zero-shot setting, as no CoT rationale examples are provided with the dataset. 

Table \ref{table:msvamp_main} presents the evaluation results on MSVAMP. MetaMath and MathOctopus were not trained on SVAMP or MSVAMP, so MSVAMP can be seen as an out-of-domain test set to evaluate domain generalization \citep{chen2023breaking}. Performance of MetaMath-\methodtable models indicates our models can generalize to out-of-domain test sets successfully.

\subsection{BBH}
Table \ref{table:bbh1} and \ref{table:bbh2} showcase the complete results for the 14 subtasks of BBH and BBH-BN. The subtasks are: \textsc{Causal Judgement, Date Understanding, Disambiguation QA, Formal Fallacies, Logical Deduction (3, 5 and 7), Navigate, Penguins in a Table, Reasoning About Colored Objects, Snarks, Sports Understanding, Temporal Sequences}, and \textsc{Web of Lies.} 

Notably, Orca 2-\methodtable models show noticeable performance degradation in \textsc{Date Understanding}. From our qualitative analysis of the CoT, we observe that Orca 2-\methodtable models frequently falsely assume an arbitrary date as the current date at the beginning of CoT (Figure \ref{fig:cot_date}), whereas the original Orca 2 models do not exhibit this behavior. Our exploration of the OpenOrca dataset reveals that examples often require the model to assume a specific current date. For example, an input text is given as \say{The current senate majority leader in the US is Chuck Schumer. Options: - yes - no}, and the target text contains \say{...Today's date is October 12, 2021...}. As we do not have access to Orca 2's original training dataset, a thorough ablation on the effect of such examples is challenging. Nevertheless, we speculate this problematic emergent behavior in Orca 2-\methodtable models is partially due to the distribution shift of the training data from the original Orca 2 dataset to the OpenOrca dataset.

\subsection{HumanEval}

\begin{table}
\scriptsize
\begin{tabularx}{\linewidth}{X}
\texttt{def greatest\_common\_divisor(a: int, b: int) -> int:}\\
\quad\quad\texttt{""" Rudi kipengele kikubwa zaidi cha pamoja cha} \\
\quad\quad\texttt{integers mbili a na b}\\
\quad\quad\texttt{>\hspace{0pt}>\hspace{0pt}> greatest\_common\_divisor(3, 5)}\\
\quad\quad\texttt{1}\\
\quad\quad\texttt{>\hspace{0pt}>\hspace{0pt}> greatest\_common\_divisor(25, 15)}\\
\quad\quad\texttt{5}\\
\quad\quad\texttt{"""} \\
\midrule
\texttt{def func(a: int, b: int) -> int:}\\
\quad\quad\texttt{""" Rudi kipengele kikubwa zaidi cha pamoja cha}\\
\quad\quad\texttt{integers mbili a na b}\\
\quad\quad\texttt{>\hspace{0pt}>\hspace{0pt}> func(3, 5)}\\
\quad\quad\texttt{1}\\
\quad\quad\texttt{>\hspace{0pt}>\hspace{0pt}> func(25, 15)}\\
\quad\quad\texttt{5}\\
\quad\quad\texttt{"""} \\
  
\end{tabularx}

\caption{
Comparison between original (\textbf{Top}) and anonymized (\textbf{Bottom}) prompts of HumanEval-MT Swahili.
}
\label{table:humaneval_comparison}
\end{table}

Table \ref{table:humaneval_comparison} compares the original and anonymized prompts of HumanEval-MT. \textbf{Top} can be solved without comprehending Swahili whereas \textbf{Bottom} is not.
Table \ref{table:humaneval_main} contains the evaluation results on the original (non-anonymized) version of HumanEval and HumanEval-MT. Compared to \method models, Code Llama models show a sharper decline in performance when evaluated on the anonymized version, suggesting that they are less capable of comprehending natural text in another language.

\section{Inference Throughput of \method Model} \label{throughput}

\begin{table}[!ht]
\small
\centering
\begin{tabularx}{1\linewidth}{l|>{\centering\arraybackslash}X>{\centering\arraybackslash}X>{\centering\arraybackslash}X}
\toprule
 & Orca 2-7B & Orca 2-\methodtable-9B & Orca 2-13B\\
\midrule

Average Time (sec) & 44.46 & 59.23 & 102.84 \\
Standard Deviation (sec) & 0.13 & 0.29 & 0.48 \\
GPU Memory (GBs) & 13.48 & 16.97 & 25.07  \\

\bottomrule
\end{tabularx}
\caption{
Inference throughputs measured using 500 examples of XCOPA.
}
\label{table:throughput}
\end{table}

We report the inference throughput of models in Table \ref{table:throughput}. We measure the total time to infer 500 instances of XCOPA. For each model, we repeat five times and report the average and the standard deviation. Additionally, we include the GPU memory utilization. We test using a single A6000 GPU on an idle server.

Results show Orca 2-LB-9B model requires considerably less compute and memory compared to Orca 2-13B, despite it outperforming Orca 2-13B in BigBenchHard Bengali (BBH-BN) and XCOPA.

\section{General-domain Language Models} \label{general_lms}

Table \ref{table:mgsm_appendix} probes the effect of \method on general-domain English-centric LMs, \textbf{Llama 2} and \textbf{Mistral 7B} \citep{jiang2023mistral}, using MGSM. Consistent with the findings on specialized LMs, \method enhances the performance of low-resource languages. As training corpora for Llama 2 and Mistral 7B are unavailable, we use a sample of the SlimPajama dataset \citep{cerebras2023slimpajama} as the training set.

\section{Ablation Studies} \label{ablations}

\subsection{Freezing/Unfreezing}

\begin{table}[ht!]
\small
\centering
\begin{tabularx}{\linewidth}{l|>{\centering\arraybackslash}s>{\centering\arraybackslash}s|>{\centering\arraybackslash}X}
\toprule
\multirow{2}{*}[-0.2em]{\textbf{Target LM}} & \multicolumn{2}{c|}{\textbf{Trainable}} & \multirow{2}{*}[-0.2em]{\textbf{AVG Score}} \\
\cmidrule(lr){2-3}
 & Enc & LM &  \\
\midrule
\multicolumn{4}{c}{\textsc{MGSM}}\\
\midrule
\multirow{2}{*}{Llama 2-7B} &  &  & 9.6 \\ 
& \Checkmark &  & \textbf{11.3} \\
\midrule
\multirow{2}{*}{Llemma-7B} & &  & 14.4 \\ 
& \Checkmark &  & \textbf{20.4} \\
\midrule
\multirow{2}{*}{MetaMath-7B} & & & \textbf{48.8} \\ 
& \Checkmark &  & 43.9 \\
\midrule
\multicolumn{4}{c}{\textsc{HumanEval + HumanEval-MT}}\\
\midrule
\multirow{2}{*}{Code Llama-7B} & & & 15.3 \\ 
& \Checkmark &  & \textbf{19.4} \\
\midrule
\multicolumn{4}{c}{\textsc{XCOPA}}\\
\midrule
\multirow{3}{*}{Orca-7B} & & & \textbf{76.6} \\ 
& \Checkmark &  & 71.1 \\
&  & \Checkmark & 74.0 \\
\midrule
\multirow{2}{*}{Orca-13B} & & & \textbf{77.3} \\ 
& \Checkmark &  & 65.1 \\
\bottomrule
\end{tabularx}
\caption{
Ablations on the effect of freezing the encoder and the LM during aligning of \method leveraging mT5-XL encoder. \Checkmark denotes the module is trainable (not frozen) during aligning.
}
\label{table:ablation_freezing}
\end{table}

We strictly keep the embedding layers of mT5 trainable throughout our experiments as we extend the vocabulary and the embedding layer to incorporate whitespace characters. mT5 tokenizers do not have whitespace characters in their vocabularies, and their default behavior is to truncate any consecutive whitespaces to a single space. However, this could negatively affect understanding code or following instructions considering the frequent use of whitespaces as delimiters (\textbackslash n, \textbackslash t, and \say{four spaces}). Therefore, even when we freeze the encoder, we leave the embedding layer trainable for the added whitespace vocabulary.  

Table \ref{table:ablation_freezing} presents the ablation study on the impact of parameter freezing during the alignment process. We apply \method with mT5-XL (2B) encoder on multiple LMs while varying the trainable modules. Notably, freezing the encoder appears beneficial when adapting finetuned LMs (MetaMath and Orca), whereas it negatively affects pretrained models (Llama, Llemma, and Code Llama). We speculate this divergence stems from differing entropy levels in the datasets: unlabeled corpora typically exhibit higher entropy than relatively well-formatted finetuning datasets. Consequently, we conjecture that for unlabeled data, keeping the encoder trainable enables the model to acclimate to the training data better. Nonetheless, we leave a thorough investigation for future research.
Additionally, training the LM during the alignment phase does not improve performance. We hypothesize this is due to the training datasets being in-domain of the LMs. As such, the LMs may not be learning additional information from the data.

\subsection{Encoder Size}

\begin{figure}[ht!]
\includegraphics[width=1\linewidth]{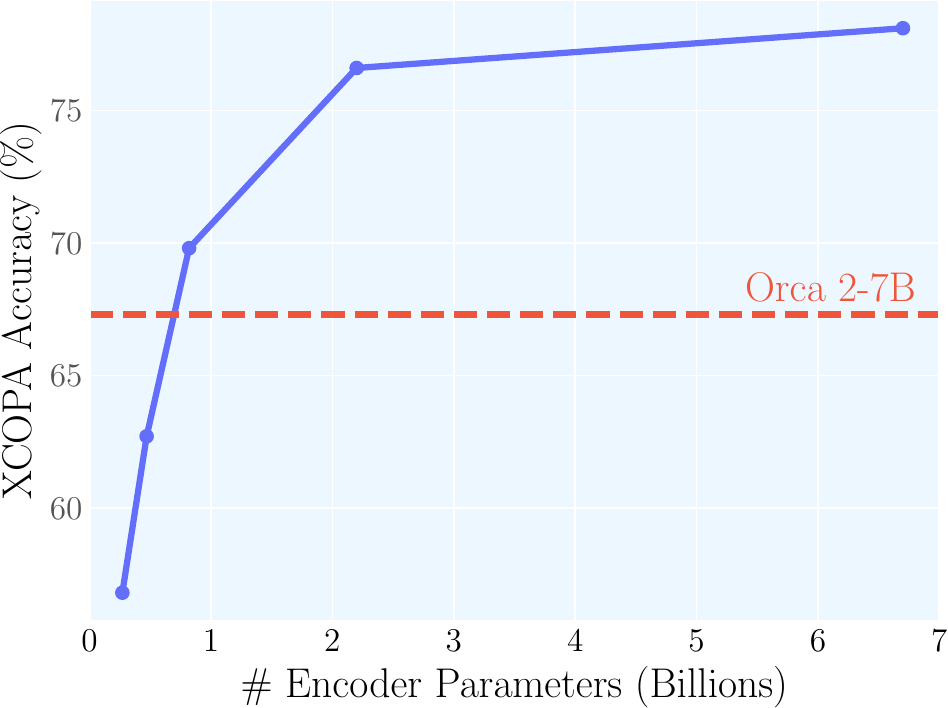}
\centering
\caption{XCOPA accuracy (\%) of Orca 2-7B models adapted with \method using five different sizes of mT5 encoder. The dotted line shows the original performance of the target LM.}
\label{fig:ablate_encodersize}
\vspace{-3mm}
\end{figure}

Figure \ref{fig:ablate_encodersize} shows the effect of encoder size on \method when applied to Orca 2-7B, measured with XCOPA. We test five different sizes of mT5 encoder: 270M (Small), 470M (Base), 820M (Large), 2.2B (XL) and 6.7B (XXL). We observe that \method with the two smaller-sized encoders underperforms the base Orca 2-7B. Nonetheless, performance increases rapidly as the encoder size scales from 270M to 2.2B and saturates in the 2.2B to 6.7B range. These results highlight that scaling the encoder size past a certain point provides diminishing returns.

\subsection{Training Set Size}

\begin{figure}[ht!]
\includegraphics[width=1\linewidth]{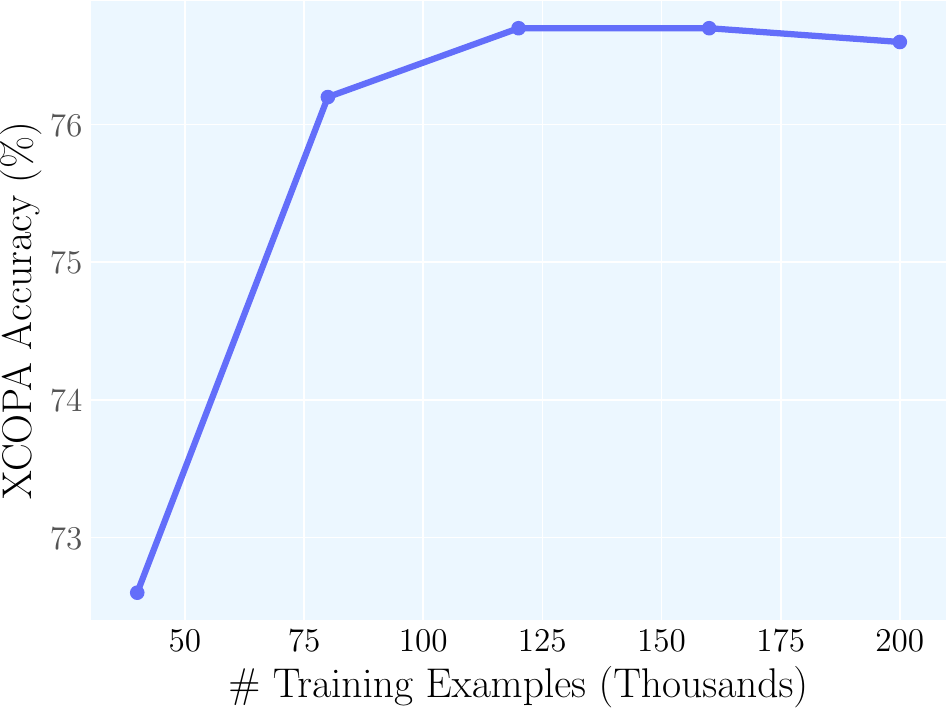}
\centering
\caption{XCOPA accuracy (\%) of Orca 2-7B models adapted with \method using five different sizes of training datasets.}
\label{fig:ablate_trainset}
\vspace{-3mm}
\end{figure}

While we fixed the training set size to 200,000 in our main experiments, Figure \ref{fig:ablate_trainset} shows that XCOPA performance peaks on 120,000 training examples, which is 60\% of our experiment setting. This suggests that in practice, \method can be applied more efficiently.

\subsection{Different Architectures}
\begin{table}[ht!]
\centering
\begin{tabularx}{0.7\linewidth}{l|>{\centering\arraybackslash}X>{\centering\arraybackslash}X}
\toprule
 & \textbf{XCOPA (Acc. \%)} \\
\midrule
Linear & 76.6 \\
MLP & 72.7 \\
Resampler & 49.7 \\
\bottomrule
\end{tabularx}
\caption{
Ablations of different \method architectures using Orca 2-7B and mT5-XL encoder.
}
\label{table:ablate_architecture}
\end{table}

Table \ref{table:ablate_architecture} shows the ablations of different \method architectures. Throughout the main experiment, we adopt a single linear layer to connect the encoder and the language model. We ablate the effect of using an MLP layer following the second iteration of LLaVa \citep{liu2023improved}, and a \say{resampler} module that adopts the architecture of perceiver resampler employed by Flamingo \citep{alayrac2022flamingo}. In contrast to the finding of \citet{liu2023improved}, we find that using an MLP layer instead of a linear layer decreases the performance slightly. Using a resampler module results in random performance.

\subsection{Different Encoder Models}

\begin{table}[ht!]
\centering
\begin{tabularx}{0.85\linewidth}{l|>{\centering\arraybackslash}X}
\toprule
 & \textbf{XCOPA (Acc. \%)} \\
\midrule
umT5-XL & 49.7 \\
umT5-XL (F) & 49.2 \\
XGLM-1.7B & 51.5 \\
XGLM-1.7B (F) & 49.7 \\
\bottomrule
\end{tabularx}
\caption{
Ablations of different encoder models for \method tested on Orca 2-7B model. (F) denotes that the encoder was frozen during alignment.
}
\label{table:ablate_encodermodel}
\end{table}

Table \ref{table:ablate_encodermodel} presents the results of using different encoder models. We test out umT5 \citep{chung2023unimax}, which improves upon mT5 to include a higher proportion of low-resource languages during pretraining. As umT5 does not have \say{LM adapted} checkpoints, unlike mT5, we use the original checkpoints. However, we find using umT5 results in random performance. Since umT5 has a nearly identical architecture to mT5, except that it has relative position bias for every Transformer layer \citep{vaswani2017attention} in contrast to the very first, we speculate that using the encoder of a non-LM Seq2Seq model resulted in failed alignment. We also test XGLM \cite{lin2022fewshot}, a decoder LM, as an encoder, in which we also observe similar results. In both cases, freezing the encoder made no difference. We leave further investigation for future research.

\section{Quality Estimation of HumanEval-MT} \label{qe_humaneval-MT}

\begin{table}[!ht]
\small
\centering
\begin{tabularx}{1\linewidth}{l|>{\centering\arraybackslash}X|>{\centering\arraybackslash}X>{\centering\arraybackslash}X>{\centering\arraybackslash}X>{\centering\arraybackslash}X>{\centering\arraybackslash}X>{\centering\arraybackslash}X}
\toprule
 & \textbf{\textsc{en}} & $ \circlearrowleft $\textbf{\textsc{sw}}& $ \circlearrowleft $\textbf{\textsc{bn}} & $ \circlearrowleft $\textbf{\textsc{pa}} & $ \circlearrowleft $\textbf{\textsc{te}} & $ \circlearrowleft $\textbf{\textsc{ur}} \\
\midrule

CL-7B & 36.0 & 35.4 & 35.4 & 36.6 & 36.6 & 37.8 \\
CL-34B & 43.9 & 36.6 & 46.3 & 42.1 & 42.1 & 45.7 \\
\midrule
BLEU & - & 88.3 & 90.5 & 93.4 & 90.5 & 96.1 \\
crhF & - & 89.9 & 93.1 & 94.6 & 93.1 & 96.5 \\

\bottomrule
\end{tabularx}
\caption{
Quality estimation of HumanEval-MT using backtranslations. CL prefix denotes Code Llama Pass@1 score.
}
\label{table:humaneval-mt_qe}
\end{table}

Table \ref{table:humaneval-mt_qe} presents the quality estimation for HumanEval-MT using backtranslation~\citep{agrawal-etal-2022-quality,zhuo-etal-2023-rethinking}. We translate HumanEval-MT back to English and measure Code Llama Pass@1 scores. As explained in Section \ref{code completion}, we anonymize all function names. In addition, we report automatic evaluation metrics, BLEU~\citep{papineni-etal-2002-bleu} and chrF~\citep{popovic-2015-chrf}, against the original HumanEval. We use the docstrings only for calculating the metrics. 

Overall, Code Llama's performance on the backtranslations matches the original HumanEval benchmark, with the exception of Code Llama-34B's performance on Swahili backtranslation. As Code Llama-7B achieves comparable results on the Swahili backtranslation, this suggests that for Swahili, GPT-4 may struggle with accurately translating complex examples that only larger models can solve. Lower automatic evaluation scores in Swahili further support the idea.

\section{Experimental Details} \label{experiemental_details}
For training \method models, we leverage AdamW \citep{loshchilov2019decoupled}. We use a learning rate of $6\mathrm{e}{-4}$ for the linear layer and $2\mathrm{e}{-5}$ for the encoder. We do not use any learning rate scheduling. We use an effective batch size of 128. 

For further training BLOOM baselines, we keep the training configuration the same as \method models with two exceptions: (1) Learning rate, in which we used a uniform value of $2\mathrm{e}{-5}$ across the entire model. (2) Sequence length, which was set to 1,152 ($1,024+128$).

For evaluations, we leverage code adapted from LM Evaluation Harness (version 0.3.0) \citep{eval-harness} and Bigcode Evaluation Harness (version 0.1.0) \citep{bigcode-evaluation-harness} packages and report single run results with fixed seed. We open source the evaluation code.

For constructing HumanEval-MT, we prompt GPT-4 with human-translated examples. The two examples used for HumanEval-MT were translated to Korean by a native Korean author. GPT-4 was then prompted to translate HumanEval to the target languages with the examples as reference. Note that the examples were provided to guide GPT-4 to keep the format of the data intact and only translate the natural language, not necessarily to enhance the translation quality itself.

\section{CoT Examples} \label{cot}
\begin{figure*}[ht!]
\includegraphics[width=0.9\linewidth]{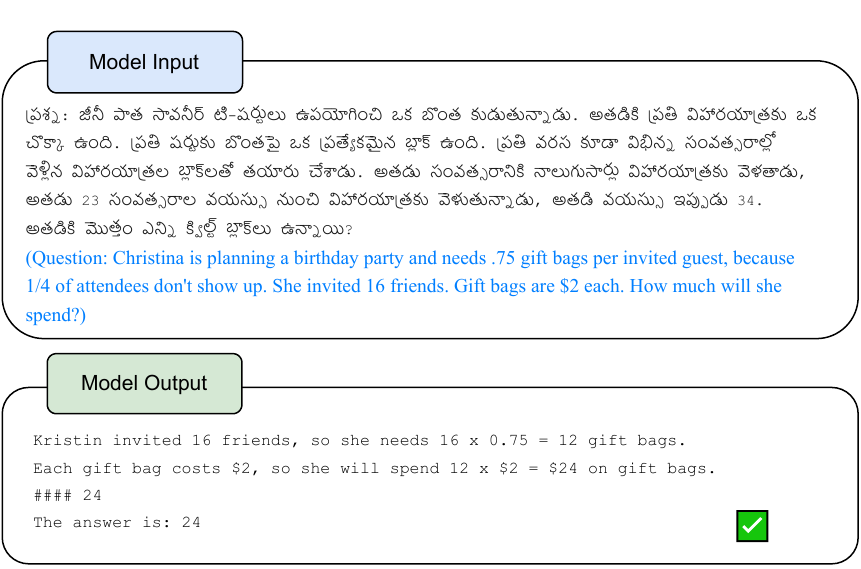}
\centering
\caption{Example of zero-shot CoT generated by MetaMath-\method-15B when prompted with an example from MGSM Telugu. For brevity, any instructional prompts are truncated. Translation of the input is provided in \blue{(blue with parenthesis)}.}
\label{fig:cot_mgsm}
\vspace{-3mm}
\end{figure*}

\begin{figure*}[ht!]
\includegraphics[width=0.9\linewidth]{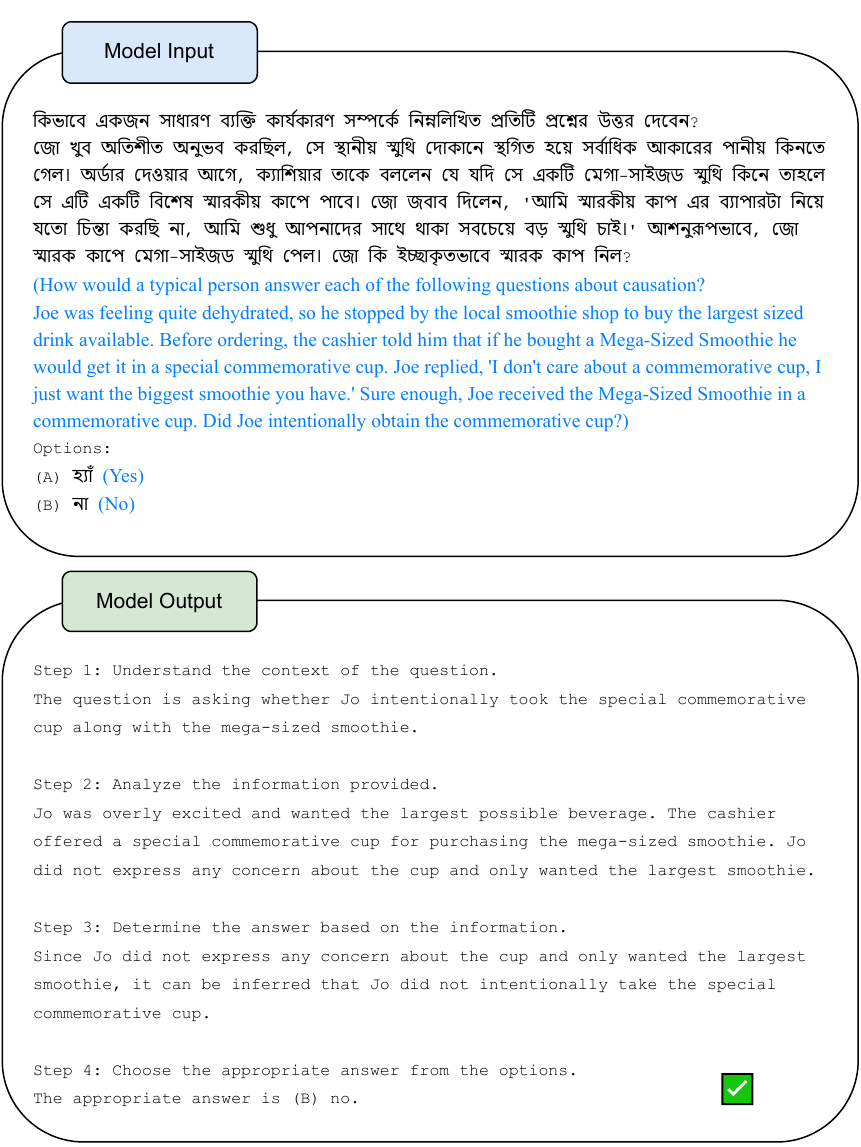}
\centering
\caption{Example of zero-shot CoT generated by Orca 2-\method-15B when prompted with an example from BBH-BN \textsc{Causal Judgement}. For brevity, any instructional prompts are truncated. Translation of the input is provided in \blue{(blue with parenthesis)}.}
\label{fig:cot_causal}
\vspace{-3mm}
\end{figure*}

\begin{figure*}[ht!]
\includegraphics[width=0.9\linewidth]{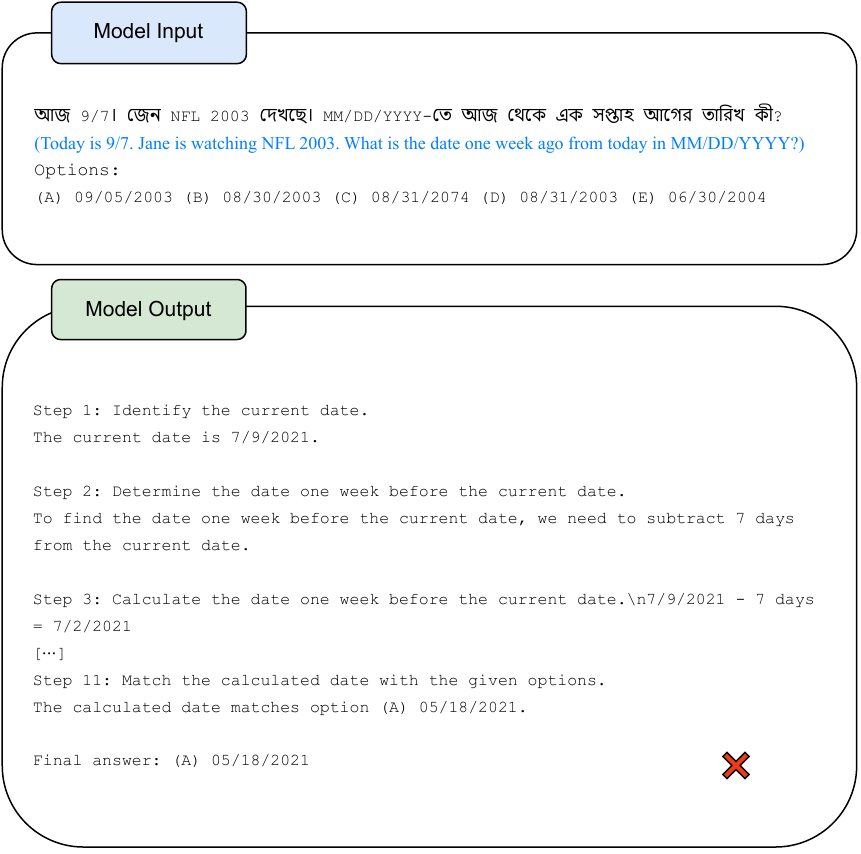}
\centering
\caption{Example of zero-shot CoT generated by Orca 2-\method-15B when prompted with an example from BBH-BN \textsc{Date Understanding}. For brevity, any instructional prompts are truncated. Translation of the input is provided in \blue{(blue with parenthesis)}.}
\label{fig:cot_date}
\vspace{-3mm}
\end{figure*}

In this section, we provide three examples of CoT generated by \method models. Figure \ref{fig:cot_mgsm} is from the MetaMath-\method-15B model prompted with an example from MGSM Telugu. Figure \ref{fig:cot_causal} and \ref{fig:cot_date} are from Orca 2-\method-15B model prompted with examples from BBH-BN \textsc{Causal Judgement} and BBH-BN \textsc{Date Understanding}. We select these two subtasks from BBH-BN as \method substantially increased the performance of Orca 2-13B for the former ($+13.4\%$), but caused a considerable decrease for the latter ($-6.4\%$). We show a success case for the former and a failure case for the other. In Appedix \ref{additional_evaluations}, we provide an explanation for the performance degradation in BBH-BN \textsc{Date Understanding}.
\begin{table*}[ht!]
\scriptsize
\centering
\begin{tabularx}{1.0\textwidth}{l|>{\centering\arraybackslash}X|>{\centering\arraybackslash}X>{\centering\arraybackslash}X>{\centering\arraybackslash}X|>{\centering\arraybackslash}X>{\centering\arraybackslash}X|>{\centering\arraybackslash}X>{\centering\arraybackslash}X>{\centering\arraybackslash}X>{\centering\arraybackslash}X>{\centering\arraybackslash}X>{\centering\arraybackslash}X>{\centering\arraybackslash}X>{\centering\arraybackslash}X>{\centering\arraybackslash}X}
\toprule
 & \textbf{\textsc{en}} & \textbf{\textsc{avg}} & \textbf{\textsc{hrl}} & \textbf{\textsc{url}} &\textbf{\textsc{zh}} & \textbf{\textsc{it}} & \textbf{\textsc{vi}} & \textbf{\textsc{id}} & \textbf{\textsc{et}} & \textbf{\textsc{ht}} & \textbf{\textsc{qu}} & \textbf{\textsc{sw}} & \textbf{\textsc{ta}} & \textbf{\textsc{th}} & \textbf{\textsc{tr}} \\
\midrule
Lang. Freq. (Llama 2, \%) & 89.7 & - & - & - & 0.13 & 0.11 & 0.08 & 0.03 & \multicolumn{7}{c}{\textsc{Less Than 0.005}}\\
\midrule
\midrule
Orca 2-7B & \textbf{98.0} & 67.3 & \textbf{86.4} & 63.0 & \textbf{85.6} & \textbf{87.2} & \textbf{83.4} & 82.4 & 54.4 & 52.2 & 49.8 & 54.2 & 58.0 & 62.0 & 71.0 \\
Orca 2-\methodtable-9B & 90.0 & \textbf{76.6} & 83.4 & \textbf{75.1} & 85.4 & 81.4 & 79.8 & \textbf{84.6} & \textbf{78.2} & \textbf{74.4} & \textbf{50.8} & \textbf{74.4} & \textbf{78.0} & \textbf{78.4} & \textbf{77.6} \\
\midrule
Orca 2-13B & \textbf{99.0} & 73.7 & \textbf{93.1} & 69.4 & \textbf{92.4} & \textbf{93.8} & \textbf{87.0} & \textbf{86.8} & 66.4 & 61.0 & 49.8 & 65.8 & 63.8 & 67.6 & 76.4 \\
Orca 2-\methodtable-15B & 92.0 & 77.3 & 84.5 & 75.7 & 85.2 & 83.8 & 83.4 & 83.8 & 80.6 & 74.8 & 50.4 & 72.2 & 77.2 & \textbf{79.8} & 79.2 \\
Orca 2-\methodtable-20B & 92.0 & \textbf{79.8} & 86.3 & \textbf{78.4} & 86.2 & 86.4 & 83.6 & 85.4 & \textbf{82.8} & \textbf{76.4} & \textbf{54.2} & \textbf{77.8} & \textbf{82.8} & \textbf{79.8} & \textbf{82.4} \\
\bottomrule
\end{tabularx}
\caption{
Accuracy (\%) on COPA and XCOPA. For XCOPA, we report the average accuracy across 11 languages.
}
\label{table:copa_main}
\end{table*}

\begin{table*}[ht!]
\small
\centering
\begin{tabularx}{1.0\textwidth}{l|>{\centering\arraybackslash}X>{\centering\arraybackslash}X>{\centering\arraybackslash}X|>{\centering\arraybackslash}X>{\centering\arraybackslash}X>{\centering\arraybackslash}X>{\centering\arraybackslash}X>{\centering\arraybackslash}X>{\centering\arraybackslash}X>{\centering\arraybackslash}X|>{\centering\arraybackslash}X>{\centering\arraybackslash}X>{\centering\arraybackslash}X}
\toprule
 & \textbf{\textsc{avg}} & \textbf{\textsc{hrl}} & \textbf{\textsc{url}} &\textbf{\textsc{en}} & \textbf{\textsc{de}} & \textbf{\textsc{fr}} & \textbf{\textsc{es}} & \textbf{\textsc{ru}} & \textbf{\textsc{zh}} & \textbf{\textsc{ja}} & \textbf{\textsc{th}} & \textbf{\textsc{sw}} & \textbf{\textsc{bn}} \\
\midrule
Lang. Freq. (Llama 2, \%) & - & - & - & 89.7 & 0.17 & 0.16 & 0.13 & 0.13 & 0.13 & 0.10 & \multicolumn{3}{c}{\textsc{Less Than 0.005}}\\
\midrule
\midrule
MathOctopus-7B & 39.2 & 41.5 & 34.0 & 39.8 & 42.4 & 44.0 & 43.3 & 41.6 & 40.4 & 38.7 & 35.1 & 29.7 & 37.2 \\
MathOctopus-13B & 45.1 & 47.2 & 40.0 & 44.8 & 46.7 & 50.6 & 49.9 & 47.6 & 47.1 & 43.9 & 40.0 & 36.3 & 43.6 \\
BLOOM-7.1B-MM & 24.9 & 29.5 & 14.1 & 40.7 & 28.8 & 33.8 & 32.0 & 27.7 & 27.7 & 15.7 & 3.0 & 21.2 & 18.0 \\
\midrule
MetaMath-7B & 47.8 & \textbf{61.1} & 16.9 & \textbf{66.3} & \textbf{63.5} & \textbf{64.1} & \textbf{64.9} & \textbf{60.3} & \textbf{55.0} & \textbf{53.4} & 19.5 & 16.8 & 14.4 \\
MetaMath-\methodtable-9B & \textbf{52.0} & 54.9 & \textbf{45.1} & 60.6 & 58.1 & 57.0 & 56.9 & 55.8 & 50.4 & 45.5 & \textbf{46.3} & \textbf{42.1} & \textbf{46.8} \\
\midrule
MetaMath-13B & 50.6 & \textbf{65.1} & 16.8 & \textbf{69.2} & \textbf{67.3} & \textbf{66.1} & \textbf{66.9} & \textbf{66.9} & \textbf{58.0} & \textbf{61.4} & 18.6 & 14.4 & 17.3 \\
MetaMath-\methodtable-15B & 57.0 & 60.4 & 49.1 & 64.1 & 61.9 & 65.9 & 64.8 & 61.0 & 55.0 & 50.1 & 51.3 & 42.1 & 54.0 \\
MetaMath-\methodtable-20B & \textbf{57.9} & 60.4 & \textbf{51.8} & 65.3 & 63.0 & 62.5 & 62.7 & 60.9 & 55.4 & 53.3 & \textbf{52.3} & \textbf{47.1} & \textbf{56.0} \\
\bottomrule
\end{tabularx}
\caption{
Accuracy (\%) on MSVAMP. MM suffix denotes the model was finetuned on MetaMath.
}
\label{table:msvamp_main}
\end{table*}

\begin{table*}[ht!]
\small
\centering
\begin{tabularx}{1.0\textwidth}{l|>{\centering\arraybackslash}X>{\centering\arraybackslash}X>{\centering\arraybackslash}X>{\centering\arraybackslash}X>{\centering\arraybackslash}X>{\centering\arraybackslash}X>{\centering\arraybackslash}X>{\centering\arraybackslash}X>{\centering\arraybackslash}X>{\centering\arraybackslash}X>{\centering\arraybackslash}X>{\centering\arraybackslash}X>{\centering\arraybackslash}X>{\centering\arraybackslash}X}
\toprule

 & \multicolumn{2}{c|}{\textsc{Causal}} & \multicolumn{2}{c|}{\textsc{Date}}   & \multicolumn{2}{c|}{\textsc{Disam.}} & \multicolumn{2}{c|}{\textsc{formal}} & \multicolumn{2}{c|}{\textsc{Logic. 3}}  & \multicolumn{2}{c|}{\textsc{Logic. 5}}  & \multicolumn{2}{c|}{\textsc{Logic. 7}}  \\
\cmidrule(lr){2-3}
\cmidrule(lr){4-5}
\cmidrule(lr){6-7}
\cmidrule(lr){8-9}
\cmidrule(lr){10-11}
\cmidrule(lr){12-13}
\cmidrule(lr){14-15}
& EN & BN & EN & BN & EN & BN & EN & BN & EN & BN & EN & BN & EN & BN\\ 
\midrule
BLOOM-7B-OpenOrca & 49.7 & 4.8 & 32.8 & 48.7 & 48.4 & 31.2 & 43.2 & 35.2 & 36.0 & 12.4 & 24.8 & 55.2 & 20.0 & 23.6 \\
\midrule
Orca 2-7B & \textbf{62.0} & 47.1 & \textbf{52.4} & \textbf{42.0} & 62.8 & 42.4 & \textbf{60.0} & 50.4 & \textbf{60.0} & 37.2 & \textbf{43.2} & 25.6 & \textbf{39.6} & 20.8  \\
Orca 2-\methodtable-9B & 57.2 & \textbf{52.9} & 26.8 & 24.4 & \textbf{64.0} & \textbf{46.8} & 55.2 & \textbf{57.2} & 52.4 & \textbf{42.0} & 36.0 & \textbf{30.0} & 38.8 & \textbf{28.8} \\
\midrule
Orca 2-13B & 56.1 & 46.5 & \textbf{64.0} & \textbf{50.0} & \textbf{66.8} & 52.0 & 52.0 & 52.0 & \textbf{68.4} & 45.2 & 46.8 & 35.6 & \textbf{49.2} & 31.2 \\
Orca 2-\methodtable-15B & 57.2 & \textbf{59.9} & 44.8 & 43.6 & 56.0 & 46.8 & \textbf{55.6} & 48.0 & 66.8 & \textbf{56.4} & \textbf{47.2} & 33.2 & 44.4 & \textbf{34.4}  \\
Orca 2-\methodtable-20B & \textbf{64.7} & 58.3 & 34.4 & 35.2 & 59.2 & \textbf{56.0} & 52.0 & \textbf{54.8} & 62.4 & 46.8 & 44.8 & \textbf{38.0} & 45.2 & 32.0 \\
\bottomrule
\end{tabularx}
\caption{
Full results on BBH and BBH-BN (Part 1 of 2).
}
\label{table:bbh1}
\end{table*}

\begin{table*}[ht!]
\small
\centering
\begin{tabularx}{1.0\textwidth}{l|>{\centering\arraybackslash}X>{\centering\arraybackslash}X>{\centering\arraybackslash}X>{\centering\arraybackslash}X>{\centering\arraybackslash}X>{\centering\arraybackslash}X>{\centering\arraybackslash}X>{\centering\arraybackslash}X>{\centering\arraybackslash}X>{\centering\arraybackslash}X>{\centering\arraybackslash}X>{\centering\arraybackslash}X>{\centering\arraybackslash}X>{\centering\arraybackslash}X}
\toprule

 & \multicolumn{2}{c|}{\textsc{Navi.}} & \multicolumn{2}{c|}{\textsc{Penguins}}   & \multicolumn{2}{c|}{\textsc{Reason.}} & \multicolumn{2}{c|}{\textsc{Snarks}} & \multicolumn{2}{c|}{\textsc{Sports}}  & \multicolumn{2}{c|}{\textsc{Tempo.}}  & \multicolumn{2}{c|}{\textsc{Web}}  \\
\cmidrule(lr){2-3}
\cmidrule(lr){4-5}
\cmidrule(lr){6-7}
\cmidrule(lr){8-9}
\cmidrule(lr){10-11}
\cmidrule(lr){12-13}
\cmidrule(lr){14-15}
& EN & BN & EN & BN & EN & BN & EN & BN & EN & BN & EN & BN & EN & BN\\ 
\midrule
BLOOM-7B-OpenOrca & 41.6 & 34.4 & 27.4 & 44.0 & 25.2 & 22.6 & 41.8 & 20.8 & 49.6 & 41.2 & 6.8 & 50.4 & 53.2 & 12.0 \\
\midrule
Orca 2-7B & \textbf{58.8} & 46.4 & \textbf{57.5} & 24.0 & \textbf{47.2} & 25.6 & \textbf{67.8} & 42.9 & \textbf{68.0} & 51.6 & \textbf{20.4} & 13.6 & \textbf{54.4} & 43.6  \\
Orca 2-\methodtable-9B & 48.8 & \textbf{50.4} & 44.5 & \textbf{39.7} & 41.6 & \textbf{33.6} & 56.5 & \textbf{53.7} & 66.4 & \textbf{53.2} & 17.6 & \textbf{17.2} & 50.8 & \textbf{54.8} \\
\midrule
Orca 2-13B & 53.2 & 49.2 & \textbf{59.6} & 30.1 & \textbf{61.6} & 26.4 & \textbf{65.5} & 48.0 & 76.4 & 49.2 & 39.6 & \textbf{22.8} & 52.0 & 45.2 \\
Orca 2-\methodtable-15B & 58.4 & \textbf{62.0} & 56.8 & \textbf{43.2} & 60.0 & 32.8 & 60.5 & 50.8 & 73.6 & 52.4 & 34.0 & 20.0 & \textbf{57.6} & 52.0 \\
Orca 2-\methodtable-20B & \textbf{60.0} & 51.2 & 50.7 & 39.0 & 59.6 & \textbf{37.2} & 62.1 & \textbf{54.8} & 72.0 & \textbf{53.6} & 24.0 & 19.2 & 52.8 & \textbf{59.2} \\
\bottomrule
\end{tabularx}
\caption{
Full results on BBH and BBH-BN (Part 2 of 2).
}
\label{table:bbh2}
\end{table*}

\begin{table*}[ht!]
\small
\centering
\begin{tabularx}{1.0\textwidth}{l|>{\centering\arraybackslash}X>{\centering\arraybackslash}X|>{\centering\arraybackslash}X>{\centering\arraybackslash}X>{\centering\arraybackslash}X>{\centering\arraybackslash}X>{\centering\arraybackslash}X>{\centering\arraybackslash}X>{\centering\arraybackslash}X>{\centering\arraybackslash}X>{\centering\arraybackslash}X>{\centering\arraybackslash}X>{\centering\arraybackslash}X>{\centering\arraybackslash}X}
\toprule

 & \multicolumn{2}{c|}{\textbf{AVG}} & \multicolumn{2}{c|}{\textbf{EN}}   & \multicolumn{2}{c|}{\textbf{SW}} & \multicolumn{2}{c|}{\textbf{BN}} & \multicolumn{2}{c|}{\textbf{PA}}  & \multicolumn{2}{c|}{\textbf{TE}}  & \multicolumn{2}{c|}{\textbf{UR}}  \\
\cmidrule(lr){2-3}
\cmidrule(lr){4-5}
\cmidrule(lr){6-7}
\cmidrule(lr){8-9}
\cmidrule(lr){10-11}
\cmidrule(lr){12-13}
\cmidrule(lr){14-15}
& & \textit{An.} & & \textit{An.} & & \textit{An.} & & \textit{An.} & & \textit{An.} & & \textit{An.} & & \textit{An.}\\ 
\midrule
Llama2-7B & 10.2 & 4.6 & 11.0 & 9.8 & 11.6 & 3.7 & 11.0 & 3.0 & 7.9 & 3.0 & 8.5 & 3.0 & 11.0 & 4.9\\
BLOOM-7.1B & 6.7 & 4.9 & 8.5 & 7.3 & 6.1 & 2.4 & 6.1 & 5.5 & 6.7 & 4.3 & 6.7 & 6.1 & 6.1 & 3.7 \\
BLOOM-7.1B-SC & 8.4 & 4.5 & 11.0 & 6.7 & 9.8 & 3.0 & 7.9 & 4.3 & 6.7 & 4.3 & 7.9 & 3.7 & 7.3 & 4.9 \\
\midrule
Code Llama-7B & 23.0 & 13.0 & \textbf{36.0} & \textbf{36.0} & 21.3 & 14.0 & 21.3 & 14.6 & 17.7 & 10.4 & 16.5 & 9.8 & \textbf{25.0} & 16.5 \\
Code Llama-\methodtable-9B & \textbf{24.9} & \textbf{19.4} & 34.2 & 31.7 & \textbf{27.4} & \textbf{17.1} & \textbf{23.2} & \textbf{15.2} & \textbf{23.2} & \textbf{18.3} & \textbf{19.5} & \textbf{15.2} & 22.0 & \textbf{18.9} \\
\midrule
Code Llama-13B & 26.0 & 19.6 & \textbf{42.7} & 40.2 & 24.4 & 15.2 & \textbf{26.2} & 17.7 & 17.7 & 12.2 & 17.7 & 12.8 & \textbf{27.4} & 19.5 \\
Code Llama-\methodtable-15B & \textbf{26.3} & \textbf{23.6} & 36.6 & \textbf{41.5} & 22.6 & 20.1 & 23.8 & \textbf{20.1} & \textbf{26.8} & 19.5 & \textbf{23.8} & \textbf{19.5} & 24.4 & \textbf{20.7} \\
Code Llama-\methodtable-20B & 26.2 & 23.5 & 35.4 & 36.0 & \textbf{25.6} & \textbf{24.4} & 22.6 & \textbf{20.1} & 25.6 & \textbf{20.7} & 20.7 & 18.9 & 27.4 & 20.7 \\
\midrule
Code Llama-34B & 29.7 & 22.7 & 46.3 & 43.9 & 28.7 & 17.1 & 32.2 & 23.8 & 22.0 & 15.9 & 19.5 & 12.8 & 29.3 & 22.6 \\
\bottomrule
\end{tabularx}
\caption{
HumanEval and HumanEval-MT Pass@1 scores. \textit{An.} denotes the anonymized version.
}
\label{table:humaneval_main}
\end{table*}

\begin{table*}[!ht]
\small
\centering
\begin{tabularx}{1\textwidth}{l|>{\centering\arraybackslash}X>{\centering\arraybackslash}X>{\centering\arraybackslash}X|>{\centering\arraybackslash}X>{\centering\arraybackslash}X>{\centering\arraybackslash}X>{\centering\arraybackslash}X>{\centering\arraybackslash}X>{\centering\arraybackslash}X>{\centering\arraybackslash}X|>{\centering\arraybackslash}X>{\centering\arraybackslash}X>{\centering\arraybackslash}X>{\centering\arraybackslash}X}
\toprule
 & \textbf{\textsc{avg}} & \textbf{\textsc{hrl}} & \textbf{\textsc{url}} &\textbf{\textsc{en}} & \textbf{\textsc{de}} & \textbf{\textsc{fr}} & \textbf{\textsc{es}} & \textbf{\textsc{ru}} & \textbf{\textsc{zh}} & \textbf{\textsc{ja}} & \textbf{\textsc{th}} & \textbf{\textsc{sw}} & \textbf{\textsc{bn}} & \textbf{\textsc{te}}\\
\midrule
Lang. Freq. (Llama 2, \%) & - & - & - & 89.7 & 0.17 & 0.16 & 0.13 & 0.13 & 0.13 & 0.10 & \multicolumn{4}{c}{\textsc{Less Than 0.005}}\\
\midrule
\midrule
Llama 2-7B & 9.1 & 12.1 & 3.9 & 15.2 & 11.6 & \textbf{13.2} & 11.2 & \textbf{11.6} & \textbf{11.2} & \textbf{10.8} & 7.2 & 5.2 & 3.2 & 0.0\\
Llama-\methodtable-9B & \textbf{11.3} & \textbf{12.2} & \textbf{9.7} & \textbf{16.8} & \textbf{12.4} & 12.8 & \textbf{13.6} & 9.2 & 10.0 & \textbf{10.8} & \textbf{13.6} & \textbf{9.2} & \textbf{7.6} & \textbf{8.4} \\ 
\midrule
Mistral-7B & \textbf{28.5} & \textbf{34.4} & 12.8 & \textbf{43.6} & \textbf{34.4} & \textbf{35.2} & \textbf{39.6} & \textbf{32.8} & \textbf{28.8} & \textbf{26.4} & 20.4 & 7.6 & 16.0 & 7.2 \\
Mistral-\methodtable-9B & 26.1 & 27.9 & \textbf{20.7} & 32.4 & 29.6 & 30.8 & 29.6 & 28.8 & 25.2 & 18.8 & \textbf{24.4} & \textbf{18.8} & \textbf{22.4} & \textbf{17.2} \\
\bottomrule
\end{tabularx}
\caption{
Accuracy (\%) of general-domain LMs on MGSM.
}
\label{table:mgsm_appendix}
\end{table*}

\end{document}